%% file: emnlp-ijcnlp-2019.tex
\documentclass[11pt,a4paper]{article}
\usepackage[hyperref]{emnlp-ijcnlp-2019}
\usepackage{times}
\usepackage{latexsym}
\usepackage{booktabs}
\usepackage{graphicx}
\usepackage{amsmath}
\usepackage{amssymb}
\usepackage{url}
\usepackage{comment}
\usepackage{color}
\usepackage{lipsum}
\newcommand\blfootnote[1]{%
  \begingroup
  \renewcommand\thefootnote{}\footnote{#1}%
  \addtocounter{footnote}{-1}%
  \endgroup
}

\aclfinalcopy % Uncomment this line for the final submission

\title{Hint-Based Training for Non-Autoregressive Machine Translation}

\author{Zhuohan Li$^{\text{1}}$\enskip Zi Lin$^{\text{2,3}}$\enskip Di He$^{\text{2}}$\enskip Fei Tian$^{\text{4}}$\enskip  Tao Qin$^{\text{5}}$\enskip Liwei Wang$^{\text{2}}$\enskip Tie-Yan Liu$^{\text{5}}$\\
$^{\text{1}}$UC Berkeley \enskip $^{\text{2}}$Peking University \enskip
$^{\text{3}}$Google AI \enskip $^{\text{4}}$Facebook \enskip $^{\text{5}}$Microsoft Research\\
{\small \tt zhuohan@cs.berkeley.edu \{zi.lin, di\_he, wanglw\}@pku.edu.cn}\\
{\small \tt feitia@fb.com \{taoqin, tyliu\}@microsoft.com}}

\date{}

\begin{document}
\maketitle\begin{abstract}

Due to the unparallelizable nature of the autoregressive factorization, AutoRegressive Translation (ART) models have to generate tokens sequentially during decoding and thus suffer from high inference latency. Non-AutoRegressive Translation (NART) models were proposed to reduce the inference time, but could only achieve inferior translation accuracy. In this paper, we proposed a novel approach to leveraging the hints from hidden states and word alignments to help the training of NART models. The results achieve significant improvement over previous NART models for the WMT14 En-De and De-En datasets and are even comparable to a strong LSTM-based ART baseline but one order of magnitude faster in inference.\blfootnote{The work was performed at Microsoft Research Asia.}
\end{abstract}

\section{Introduction}

Neural machine translation has attracted much attention in recent years \citep{bahdanau2014neural, bahdanau2016actor, kalchbrenner2016neural, gehring2016convolutional}. Given a sentence $x=(x_1, \dots,x_{T_x})$ from the source language, the straight-forward way for translation is to generate the target words $y=(y_1, \dots, y_{T_y})$ one by one from left to right. This is also known as the \emph{AutoRegressive} Translation (ART) models, in which the joint probability is decomposed into a chain of conditional probabilities:
\begin{eqnarray}
\label{eq:fac}
P(y|x)=\Pi^{T_y}_{t=1} P(y_t|y_{<t},x),
\end{eqnarray}

While the ART models have achieved great success in terms of translation quality, the time consumption during inference is still far away from satisfactory. During training, the predictions at different positions can be estimated in parallel since the ground truth pair $(x,y)$ is exposed to the model. However, during inference, the model has to generate tokens sequentially as $y_{<t}$ must be inferred on the fly. Such autoregressive behavior becomes the bottleneck of the computational time \citep{wu2016google}.  

In order to speed up the inference process, a line of works begin to develop non-autoregressive translation models. These models break the autoregressive dependency by decomposing the joint probability with
\begin{equation}
P(y|x)=P(T_y|x)\Pi^{T_y}_{t=1} P(y_t|x). 
\end{equation}

The lost of autoregressive dependency largely hurt the consistency of the output sentences, increase the difficulty in the learning process and thus lead to a low quality translation. Previous works mainly focus on adding different components into the NART model to improve the expressiveness of the network structure to overcome the loss of autoregressive dependency \citep{gu2017non, lee2018deterministic, kaiser2018fast}. However, the computational overhead of new components will hurt the inference speed, contradicting with the goal of the NART models: to parallelize and speed up neural machine translation models.

To tackle this, we proposed a novel hint-based method for NART model training. We first investigate the causes of the poor performance of the NART model. Comparing with the ART model, we find that: (1) the positions where the NART model outputs incoherent tokens will have very high hidden states similarity; (2) the attention distributions of the NART model are more ambiguous than those of ART model. Therefore, we design two kinds of hints from the hidden states and attention distributions of the ART model to help the training of the NART model. The experimental results show that our model achieves significant improvement over the NART baseline models and is even comparable to a strong ART baseline in \citet{wu2016google}.

\begin{figure*}[tb]
    \centering
  \centerline{\includegraphics[width = 14cm]{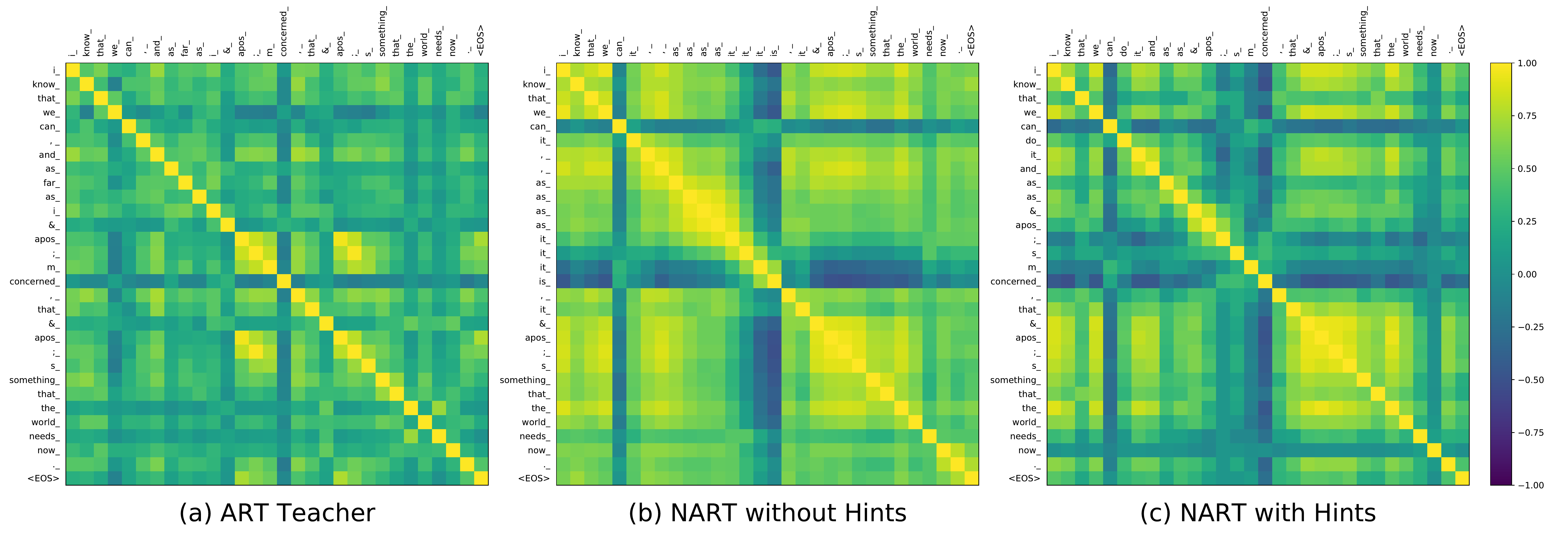}}
    \caption{Case study: the above three figures visualize the hidden state cosine similarities of different models. The axes correspond to the generated target tokens. Each pixel shows the cosine similarities $\cos_{ij}$ between the last layer hidden states of the $i$-th and $j$-th generated tokens, where the diagonal pixel will always be 1.0. }
    \label{fig:hidden}
\end{figure*}

\section{Approach}
In this section, we first describe the observations on the ART and NART models, and then discuss what kinds of information can be used as hints to help the training of the NART model. We follow the network structure in \citet{vaswani2017attention}, use a copy of the source sentence as decoder input, remove the attention masks in decoder self-attention layers and add a positional attention layer as suggested in \citet{gu2017non}. We provide a visualization of ART and NART models we used in Figure \ref{fig:transformer} and a detailed description of the model structure in Appendix.

\subsection{Observation: Illed States and Attentions}
According to the case study in \citet{gu2017non}, the translations of the NART models contain incoherent phrases (e.g. repetitive words) and miss meaningful tokens on the source side, while these patterns do not commonly appear in ART models. After some empirical study, we find two non-obvious facts that lead to this phenomenon.

First, we visualize the cosine similarities between decoder hidden states of a certain layer in both ART and NART models for sampled cases. Mathematically, for a set of hidden states $r_1, \ldots, r_T$, the pairwise cosine similarity can be derived by $\cos_{ij} = {\left<r_i, r_j\right>}/{(\|r_i\|\cdot\|r_j\|)}.$ We then plot the heatmap of the resulting matrix $\cos$. A typical example is shown in Figure~\ref{fig:hidden}, where the cosine similarities in the NART model are larger than those of the ART model, indicating that the hidden states across positions in the NART model are ``similar''. Positions with highly-correlated hidden states tend to generate the same word and make the NART model output repetitive tokens, e.g., the yellow area on the top-left of Figure~\ref{fig:hidden}(b), while this does not happen in the ART model (Figure~\ref{fig:hidden}(a)). According to our statistics, 70\% of the cosine similarities between hidden states in the ART model are less than 0.25, and 95\% are less than 0.5.

\begin{figure*}[tb]
    \centering
  \centerline{\includegraphics[width = 14cm]{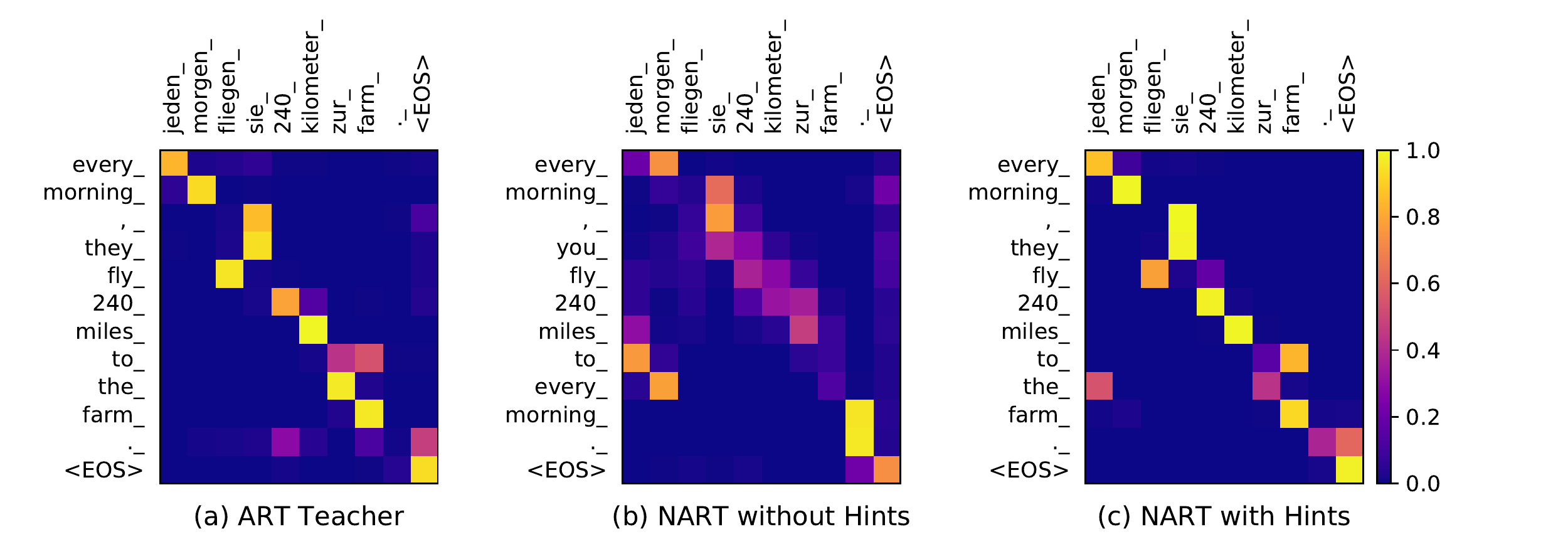}}
    \caption{Case study: the above three figures visualize the encoder-decoder attention weights of different models. The x-axis and y-axis correspond to the source and generated target tokens respectively. The attention distribution is from a single head of the third layer encoder-decoder attention, which is the most informative one according to our observation. Each pixel shows attention weights $\alpha_{ij}$ between the $i$-th source token and $j$-th target token. }
    \label{fig:attention}
\end{figure*}

Second, we visualize the encoder-decoder attentions for sampled cases, shown in Figure \ref{fig:attention}. Good attentions between the source and target sentences are usually considered to lead to accurate translation while poor ones may cause wrong output tokens \citep{bahdanau2014neural}. In Figure~\ref{fig:attention}(b), the attentions of the ART model almost covers all source tokens, while the attentions of the NART model do not cover ``farm'' but with two ``morning''. This directly makes the translation result worse in the NART model. These phenomena inspire us to use the intermediate hidden information in the ART model to guide the learning process of the NART model.

\subsection{Hints from the ART teacher Model}

Our study motivates us to leverage the intermediate hidden information from an ART model to improve the NART model. We focus on how to define hints from a well-trained ART \emph{teacher} model and use it to guide the training process of a NART \emph{student} model. We study layer-to-layer hints and assume both the teacher and student models have an $M$-layer encoder and an $N$-layer decoder, despite the difference in stacked components. 

Without the loss of generality, we discuss our method on a given paired sentence $(x,y)$. In real experiments, losses are averaged over all training data. For the teacher model, we use $a_{t,l,h}^\mathit{tr}$ as the encoder-to-decoder attention distribution of $h$-th head in the $l$-th decoder layer at position $t$, and use $r_{t,l}^\mathit{tr}$ as the output of the $l$-th decoder layer after feed forward network at position $t$. Correspondingly, $a_{t,l,h}^\mathit{st}$ and $r_{t,l}^\mathit{st}$ are used for the student model. We propose a hint-based training framework that contains two kinds of hints:

\paragraph{Hints from hidden states}  
The discrepancy of hidden states motivates us to use hidden states of the ART model as a hint for the learning process of the NART model. One straight-forward method is to regularize the $L_1$ or $L_2$ distance between each pair of hidden states in ART and NART models. However, since the network components are quite different in ART and NART models, applying the straight-forward regression on hidden states hurts the learning process and fails. Therefore, we design a more implicit loss to help the student refrain from the incoherent translation results by acting towards the teacher in the hidden-state level:
\begin{align*}
\mathcal{L}_\mathit{hid} = {}& \frac{2}{(T_y-1)T_yN} \sum_{s=1}^{T_y-1}\sum_{t=s+1}^{T_y}\sum_{l=1}^{N} \phi(d_\mathit{st}, d_\mathit{tr}),
\end{align*}

where $d_\mathit{st} = \cos(r_{s, l}^\mathit{st},r_{t, l}^\mathit{st})$, $d_\mathit{tr} = \cos(r_{s, l}^\mathit{tr},r_{t, l}^\mathit{tr})$, and $\phi$ is a penalty function. In particular, we let 
\begin{align*}
\phi(d_\mathit{st}, d_\mathit{tr}) = \begin{cases}
-\log(1 - d_\mathit{st}), & \text{if } d_\mathit{st} \geq \gamma_\mathit{st} \\
& \text{and } d_\mathit{tr} \leq \gamma_\mathit{tr}; \\
0, & \text{else,}
\end{cases}
\end{align*}
where $-1\leq\gamma_\mathit{st}, \gamma_\mathit{tr}\leq1$ are two thresholds controlling whether to penalize or not. We design this loss since we only want to penalize hidden states that are highly similar in the NART model, but not similar in the ART model. We have tested several choices of $-\log(1-d_\mathit{st})$, e.g., $\exp(d_\mathit{st})$, from which we find similar experimental results.

\paragraph{Hints from word alignments} 
We observe that meaningful words in the source sentence are sometimes untranslated by the NART model, and the corresponding positions often suffer from ambiguous attention distributions. Therefore, we use the word alignment information from the ART model to help the training of the NART model.

In particular, we minimize KL-divergence between the per-head encoder-to-decoder attention distributions of the teacher and the student to encourage the student to have similar word alignments to the teacher model, i.e.
\begin{align*}
\mathcal{L}_\mathit{align}
={}& \frac{1}{T_yNH} \sum_{t=1}^{T_y}\sum_{l=1}^{N}\sum_{h=1}^{H} D_\text{KL}(a^\mathit{tr}_{t, l, h}\| a^\mathit{st}_{t, l, h}). 
\end{align*}

Our final training loss $\mathcal{L}$ is a weighted sum of two parts stated above and the negative log-likelihood loss $\mathcal{L}_\mathit{nll}$ defined on bilingual sentence pair $(x, y)$, i.e.
\begin{eqnarray}
\mathcal{L} = \mathcal{L}_\mathit{nll} + \lambda \mathcal{L}_\mathit{hid} + \mu\mathcal{L}_\mathit{align},
\end{eqnarray}
where $\lambda$ and $\mu$ are hyperparameters controlling the weight of different loss terms.

\begin{figure*}[tb]
    \centering
	\centerline{\includegraphics[width = 12.7cm]{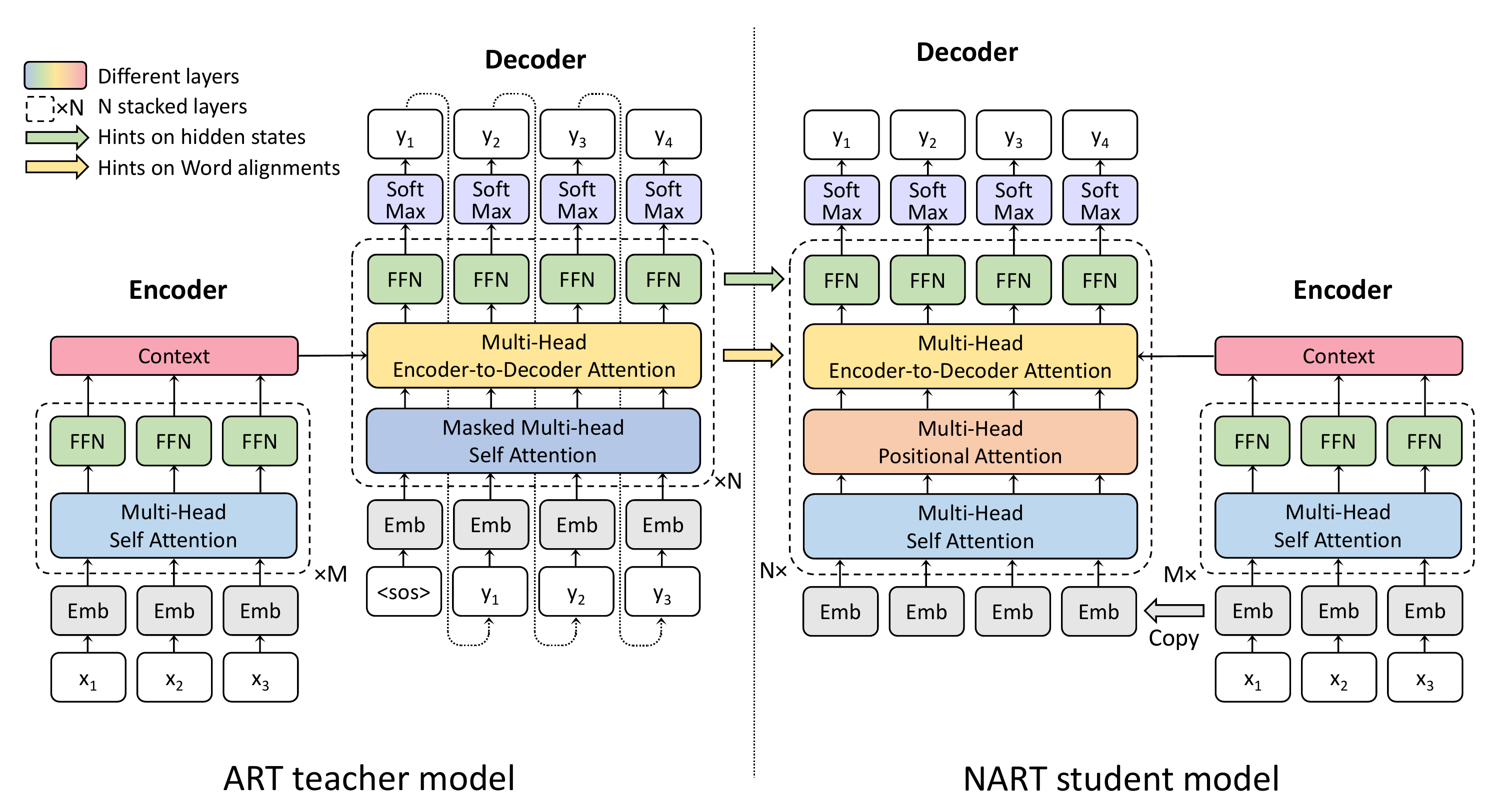}}
    \caption{Hint-based training from ART model to NART model.}
    \label{fig:transformer}
\end{figure*}

\section{Experiments}
\label{sec:exp}
\subsection{Experimental Settings}

The evaluation is on two widely used public machine translation datasets: IWSLT14 German-to-English (De-En) \citep{huang2017neural,bahdanau2016actor}  and WMT14 English-to-German (En-De) dataset \citep{wu2016google,gehring2017convolutional}. To compare with previous works, we also reverse WMT14 English-to-German dataset and obtain WMT14 German-to-English dataset.

We pretrain Transformer \citep{vaswani2017attention} as the teacher model on each dataset, which achieves 33.26/27.30/31.29 in terms of BLEU \citep{papineni2002bleu} in IWSLT14 De-En, WMT14 En-De and De-En test sets. The student model shares the same number of layers in encoder/decoder, size of hidden states/embeddings and number of heads as the teacher models (Figure~\ref{fig:transformer}). Following \citet{gu2017non, kim2016sequence}, we replace the target sentences by the decoded output of the teacher models.

Hyperparameters ($\gamma_\mathit{st}, \gamma_\mathit{tr}, \lambda, \mu$) for hint-based learning  are determined to make the scales of three loss components similar after initialization. We also employ label smoothing of value $\epsilon_\mathit{ls}=0.1$ \citep{szegedy2016rethinking} in all experiments. We use Adam optimizer and follow the setting in \citet{vaswani2017attention}. Models for WMT14/IWSLT14 tasks are trained on 8/1 NVIDIA M40 GPUs respectively. The model is based on the open-sourced \texttt{tensor2tensor} \citep{tensor2tensor}.\footnote{Open-source code can be found at \url{https://github.com/zhuohan123/hint-nart}} More settings can be found in Appendix.

\subsection{Inference}
During training, $T_y$ does not need to be predicted as the target sentence is given. During testing, we have to predict the length of the target sentence for each source sentence. In many languages, the length of the target sentence can be roughly estimated from the length of the source sentence. We choose a simple method to avoid the computational overhead, which uses input length to determine target sentence length: $T_y = T_x + C$, where $C$ is a constant bias determined by the average length differences between the source and target training sentences. We can also predict the target length ranging from $[(T_x+C)-B, (T_x+C)+B]$, where $B$ is the halfwidth. By doing this, we can obtain multiple translation results with different lengths. Note that we choose this method only to show the effectiveness of our proposed method and a more advanced length estimation method can be used to further improve the performance.

Once we have multiple translation results, we additionally use our ART teacher model to evaluate each result and select the one that achieves the highest probability. As the evaluation is fully parallelizable (since it is identical to the parallel training of the ART model), this rescoring operation will not hurt the non-autoregressive property of the NART model. 

\begin{table*}
\centering
\centerline{
\scalebox{0.75}{
\begin{tabular}{l|ccc|cc}
\toprule
                & \multicolumn{2}{c}{\textbf{WMT14}}    & \textbf{IWSLT14}      &                   & \\ 
\textbf{Models} & En-De            & De-En              & De-En                 & \textbf{Latency}  &  \textbf{Speedup}\\
\midrule
\multicolumn{2}{l}{\textit{Autoregressive models}} \\
\midrule
LSTM-based S2S \citep{wu2016google, bahdanau2016actor}     & 24.60        & /             & 28.53        & / & /\\
ConvS2S \citep{gehring2017convolutional, edunov2017classical} & 26.43        & /             & 32.84        & / & /\\
Transformer  \citep{vaswani2017attention}   & 27.30 & 31.29  & 33.26  & 784 ms$^\ddag$& 1.00$\times$\\
\midrule
\multicolumn{2}{l}{\textit{Non-autoregressive models}} \\
\midrule
FT \citep{gu2017non}                     & 17.69 & 20.62         & /     & 39 ms$^\dag$ & 15.6$\times$\\
FT (rescoring 10 candidates)                & 18.66 & 22.41         & /     & 79 ms$^\dag$ & 7.68$\times$\\
FT (rescoring 100 candidates)                & 19.17 & 23.20         & /     & 257 ms$^\dag$ & 2.36$\times$\\
IR \citep[adaptive refinement steps]{lee2018deterministic}     & 21.54 & 25.43         & /     & / & 2.39$\times$ \\
LT \citep{kaiser2018fast}                      & 19.8  & /             & /     & 105 ms$^\dag$ & 5.78$\times$\\
LT (rescoring 10 candidates)            & 21.0  & /             & /     & / & / \\
LT (rescoring 100 candidates)           & 22.5  & /             & /     & / & / \\
\textbf{NART w/ hints}                     & \textbf{21.11} &  \textbf{25.24}   & \textbf{25.55} & \textbf{26 ms}$^\ddag$& \textbf{30.2$\boldsymbol{\times}$}\\
\textbf{NART w/ hints} ($B = 4$, 9 candidates)& \textbf{25.20} & \textbf{29.52} & \textbf{28.80} & \textbf{44 ms}$^\ddag$ & \textbf{17.8}$\boldsymbol{\times}$\\
\bottomrule
\end{tabular}}
}
\caption{Performance on WMT14 En-De, De-En and IWSLT14 De-En tasks. ``/'' means non-reportable. }
\label{tbl:exp_result}
\end{table*}

\subsection{Experimental Results}

We compare our model with several baselines, including three ART models, the fertility based (FT) NART model \citep{gu2017non}, the deterministic iterative refinement based (IR) NART model \citep{lee2018deterministic}, and the Latent Transformer \citep[LT;][]{kaiser2018fast} which is not fully non-autoregressive by incorporating an autoregressive sub-module in the NART model architecture.

The results are shown in the Table~\ref{tbl:exp_result}.\footnote{$\ddag$ and $\dag$ indicate that the latency is measured on our own platform or by previous works, respectively. Note that the latencies may be affected by hardware settings and such absolute values are not fair for direct comparison, so we also list the speedup of the works compared to their ART baselines.} Across different datasets, our method achieves significant improvements over previous non-autoregressive models. Specifically, our method outperforms fertility based NART model with 6.54/7.11 BLEU score improvements on WMT En-De and De-En tasks in similar settings and achieves comparable results with state-of-the-art LSTM-based model on WMT En-De task. Furthermore, our model achieves a speedup of 30.2 (output a single sentence) or 17.8 (teacher rescoring) times over the ART counterparts. Note that our speedups significantly outperform all previous works, because of our lighter design of the NART model: without any computationally expensive module trying to improve the expressiveness.

We also visualize the hidden state cosine similarities and attention distributions for the NART model with hint-based training, as shown in Figure~\ref{fig:hidden}(c) and \ref{fig:attention}(c). With hints from hidden states, the hidden states similarities of the NART model decrease in general, and especially for the positions where the original NART model outputs incoherent phrases. The attention distribution of the NART model after hint-based training is more similar to the ART teacher model and less ambiguous comparing to the NART model without hints.

According to our empirical analysis, the percentage of repetitive words drops from 8.3\% to 6.5\% by our proposed methods on the IWSLT14 De-En test set, which is a 20\%+ reduction. This shows that our proposed method effectively improve the quality of the translation outputs. We also provide several case studies in Appendix.

\begin{table}[ht]
  \centering
\scalebox{0.66}{
  \begin{tabular}{c|ccc}
    \toprule
    \textbf{Model} & $\mathcal{L}_\mathit{nll}$ & $\mathcal{L}_\mathit{nll}+\mathcal{L}_\mathit{align}$ & $\mathcal{L}_\mathit{nll}+\mathcal{L}_\mathit{align}+\mathcal{L}_\mathit{hid}$ \\
    \midrule
    \textbf{BLEU} & 23.08 & 24.76 & \textbf{25.55}\\
    \textbf{Long-sentence BLEU} & 17.48 & 19.24 & \textbf{20.63}\\
    \bottomrule
  \end{tabular}
}
  \caption{Ablation studies on IWSLT14 De-En. Results are BLEU scores without teacher rescoring.}
  \label{tbl:ablation}
\end{table}

Finally, we conduct an ablation study on IWSLT14 De-En task. As shown in Table~\ref{tbl:ablation}, the hints from word alignments provide an improvement of about 1.6 BLEU points, and the hints from hidden states improve the results by about 0.8 BLEU points. We also test these models on a subsampled set whose source sentence lengths are at least 40. Our model outperforms the baseline model by more than 3 BLEU points (20.63 v.s. 17.48).

\section{Conclusion}

In this paper, we proposed to use hints from a well-trained ART model to enhance the training of NART models. Our results on WMT14 En-De and De-En significantly outperform previous NART baselines, with one order of magnitude faster in inference than ART models. In the future, we will focus on designing new architectures and training methods for NART models to achieve comparable accuracy as ART models.

\section*{Acknowledgment}
This work is supported by National Key R\&D Program of China (2018YFB1402600),
NSFC (61573026) and BJNSF (L172037) and a grant from
Microsoft Research Asia. We would like to thank the anonymous reviewers for their valuable comments on our paper.

\bibliography{emnlp-ijcnlp-2019}
\bibliographystyle{acl_natbib}
\input{emnlp-ijcnlp-appendix.tex}
\end{document}

%% file: emnlp-ijcnlp-appendix.tex
\newpage
\appendix
\section*{Appendix}

\section{Related Works}

\subsection{AutoRegressive Translation}

Given a sentence $x=(x_1, \dots,x_{T_x})$ from the source language, the straight-forward way for translation is to generate the words in the target language $y=(y_1, \dots, y_{T_y})$ one by one from left to right. This is also known as the autoregressive factorization in which the joint probability is decomposed into a chain of conditional probabilities, as in the Eqn.~(\ref{eq:fac}). Deep neural networks are widely used to model such conditional probabilities based on the encoder-decoder framework. The encoder takes the source tokens $(x_1, \dots,x_{T_x})$ as input and encodes $x$ into a set of context states $c=(c_1,\dots,c_{T_x})$. The decoder takes $c$ and subsequence $y_{<t}$ as input and estimates $P(y_t|y_{<t},c)$ according to some parametric function. 

There are many design choices in the encoder-decoder framework based on different types of layers, e.g., recurrent neural network(RNN)-based \citep{bahdanau2014neural}, convolution neural network(CNN)-based \citep{gehring2017convolutional} and recent self-attention based \citep{vaswani2017attention} approaches. We show a self-attention based network (Transformer) in the left part of Figure~\ref{fig:transformer}. While the ART models have achieved great success in terms of translation quality, the time consumption during inference is still far away from satisfactory. During training, the ground truth pair $(x,y)$ is exposed to the model, and thus the prediction at different positions can be estimated in parallel based on CNN or self-attention networks. However, during inference, given a source sentence $x$, the decoder has to generate tokens sequentially, as the decoder inputs $y_{<t}$ must be inferred on the fly. Such autoregressive behavior becomes the bottleneck of the computational time \citep{wu2016google}. 

\subsection{Non-AutoRegressive Translation}
\label{sec:nart}
In order to speed up the inference process, a line of works begin to develop non-autoregressive translation models. These models follow the encoder-decoder framework and inherit the encoder structure from the autoregressive models. After generating the context states $c$ by the encoder, a separate module will be used to predict the target sentence length $T_y$ and decoder inputs $z = (z_1, \ldots, z_{T_y})$ by a parametric function: $(T_y, z) \sim f_z(x, c;\theta)$, which is either deterministic or stochastic. The decoder will then predict $y$ based on following probabilistic decomposition
\begin{eqnarray}
P(y|x, T_y, z)=\Pi^{T_y}_{t=1} P(y_t|z, c). \label{eq:nat_decomposition}
\end{eqnarray}
Different configurations of $T_y$ and $z$ enable the decoder to produce different target sentence $y$ given the same input sentence $x$, which increases the output diversity of the translation models. 

Previous works mainly pay attention to different design choices of $f_z$. \citet{gu2017non} introduce \emph{fertilities}, corresponding to the number of target tokens occupied by each of the source tokens, and use a non-uniform copy of encoder inputs as $z$ according to the fertility of each input token. The prediction of fertilities is done by a separated neural network-based module. \citet{lee2018deterministic} define $z$ by a sequence of generated target sentences $y^{(0)}, \ldots, y^{(L)}$, where each $y^{(i)}$ is a refinement of $y^{(i-1)}$. \citet{kaiser2018fast} use a sequence of autoregressively generated discrete latent variables as inputs of the decoder.

While the expressiveness of $z$ improved by different kinds of design choices, the computational overhead of $z$ will hurt the inference speed of the NART models. Comparing to the more than 15$\times$ speed up in \citet{gu2017non}, which uses a relatively simpler design choice of $z$, the speedup of \citet{kaiser2018fast} is reduced to about 5$\times$, and the speedup of \citet{lee2018deterministic} is reduced to about 2$\times$. This contradicts with the design goal of the NART models: to parallelize and speed up neural machine translation models. 

\subsection{Knowledge Distillation and Hint-Based Training}

Knowledge Distillation (KD) was first proposed by \citet{hinton2015distilling}, which trains a small \emph{student network} from a large (possibly ensemble) \emph{teacher network}. The training objective of the student network contains two parts. The first part is the standard classification loss, e.g, the cross entropy loss defined on the student network and the training data. The second part is defined between the output distributions of the student network and the teacher network, e.g, using KL-divergence . \citet{kim2016sequence} introduces the KD framework to neural machine translation models. They replace the ground truth target sentence by the generated sentence from a well-trained teacher model. Sentence-level KD is also proved helpful for non-autoregressive translation in multiple previous works \citep{gu2017non, lee2018deterministic}.

However, knowledge distillation only uses the outputs of the teacher model, but ignores the rich hidden information inside a teacher model. \citet{romero2014fitnets} introduced \emph{hint-based training} to leverage the intermediate representations learned by the teacher model as hints to improve the training process and final performance of the student model. \citet{hu2018attention} used the attention weights as hints to train a small student network for reading comprehension.

\section{Network Architecture}
\label{network-structure}
\paragraph{Encoder and decoder}
Same as the ART model, the encoder of the NART model takes the embeddings of source tokens as inputs,\footnote{Following \cite{vaswani2017attention,gu2017non} we also use \emph{positional embedding} to model relative correlation between positions and add it to word embedding in both source and target sides. The positional embedding is represented by a sinusoidal function of different frequencies to encode different positions. Specifically, the positional encoding $e_\mathit{pos}$ is computed as $e_\mathit{pos}(j, k) = \sin(j/10000^{k/d})$ (for even $k$) or $\cos(j/10000^{k/d})$ (for odd $k$), where $j$ is the position index and $k$ is the dimension index of the embedding vector. 
} and generates a set of context vectors. As discussed in the main paper, the NART model needs to predict $z$ given length $T_y$ and source sentence $x$. We use a simple and efficient method to predict $z=(z_1, \dots,z_{T_y})$. Given source sentence $x = (x_1, \ldots, x_{T_x})$ and target length $T_y$, we denote $e(x_i)$ as the embedding of $x_i$. We linearly combine the embeddings of all the source tokens to generate $z$ as follows:
\begin{eqnarray}
z_j = \sum_{i}w_{ij} \cdot e(x_i),
\end{eqnarray}
where $w_{ij}$ is the normalized weight that controls the contribution of $e(x_i)$ to $z_j$ according to
\begin{eqnarray}
w_{ij} {}\propto{} \exp\left(-{(j-j'(i))^2}/{\tau}\right),
\end{eqnarray}
where $j=1,\ldots,T_y$ and $j'(i) = (T_y/T_x)\cdot i$. $\tau$ is a hyperparameter to control the ``sharpness'' of the weight distribution. We use $f_z(x,T_y,\tau)$ for this weighted average function to be consistent as in the non-autoregressive decomposition.

\paragraph{Three types of multi-head attention} The ART and NART models share two types of multi-head attentions: multi-head \emph{self} attention and multi-head \emph{encoder-to-decoder} attention. The NART model specifically uses multi-head \emph{positional} attention to model local word orders within the sentence \citep{vaswani2017attention,gu2017non}. A general attention mechanism can be formulated as querying a dictionary with key-value pairs \citep{vaswani2017attention}, e.g.,
\begin{align}
&\text{Attention}(Q,K,V)\notag\\={}&\text{softmax}\left(\frac{QK^T}{\sqrt{d_{model}}}\right)\cdot V,
\end{align}
where $d_\mathit{model}$ is the dimension of hidden representations and $Q$ (Query), $K$ (Key), $V$ (Value) differ among three types of attentions. For self attention, $Q$, $K$ and $V$ are hidden representations of the previous layer. %\footnote{In ART decoder, self-attention at each position is used with masking out all subsequent positions, in order to allow the attention only computed over its preceding positions \cite{vaswani2017attention}. Since there is no such inherent dependency in NART decoder, we mask out each query position only to avoid from attending to itself following \cite{gu2017non}. } 
 For encoder-to-decoder attention, $Q$ is hidden representations of the previous layer, whereas $K$ and $V$ are context vectors from the encoder. For positional attention, positional embeddings are used as $Q$ and $K$, and hidden representations of the previous layer are used as $V$. The multi-head variant of attention allows the model to jointly attend to information from different representation subspaces, and is defined as
\begin{align}
&\text{Multi-head}(Q,K,V) \notag\\ ={}& \text{Concat} (\text{head}_1,\cdots,\text{head}_H)W^O,\\
&\text{head}_h\notag\\ ={}& \text{Attention}(QW_h^Q, KW_h^K,VW_h^V),
\end{align}
where $W_h^Q\in \mathbb{R}^{d_\mathit{model}\times d_k}$, $W_h^K\in \mathbb{R}^{d_\mathit{model}\times d_k}$, $W_h^V\in \mathbb{R}^{d_\mathit{model}\times d_k},$ and $W_h^O\in \mathbb{R}^{d_\mathit{model}\times Hd_v}$ are project parameter matrices, $H$ is the number of heads, and $d_k$ and $d_v$ are the numbers of dimensions. 

In addition to multi-head attentions, the encoder and decoder also contain fully connected feed-forward network (FFN) layers with ReLU activations, which are applied to each position separately and identically. Compositions of self attention, encoder-to-decoder attention, positional attention, and position-wise feed-forward network are stacked to form the encoder and decoder of the ART model and the NART model, with residual connections \citep{he2016deep} and layer normalization \citep{ba2016layer}.

\begin{table*}[t]
    \centering
	\centerline{\includegraphics[width = 14cm]{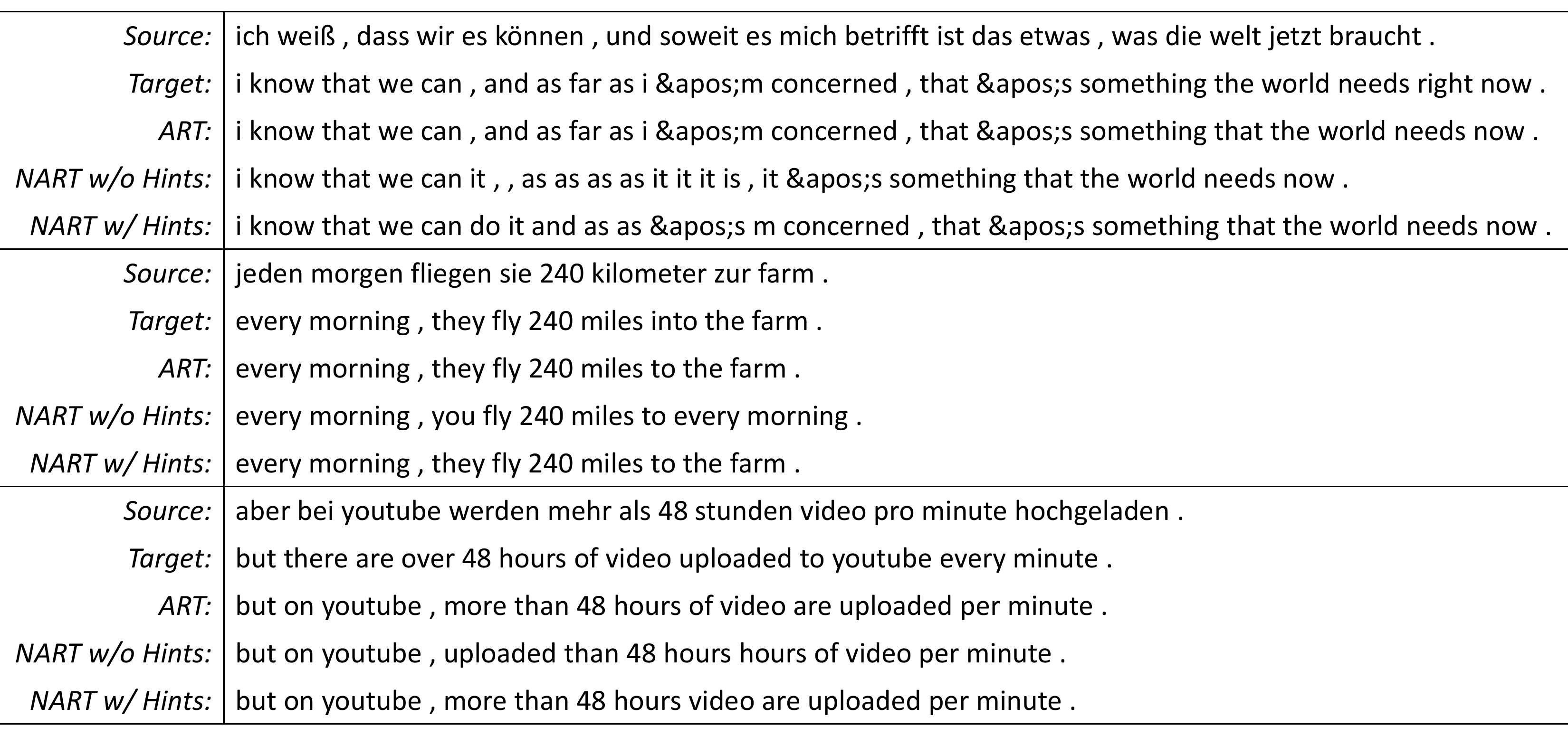}}
	\caption{Cases on IWSLT14 De-En.}
    \label{exp:case-study}
\end{table*}

\section{Extra Experimental Settings}
\label{sec:extraexp}
\paragraph{Dataset specifications}

The split of the training/validation/test sets of the IWSLT14 dataset\footnote{https://wit3.fbk.eu/} contain about 153K/7K/7K sentence pairs, respectively. The training set of the WMT14 dataset\footnote{http://www.statmt.org/wmt14/translation-task} contains 4.5M parallel sentence pairs. Newstest2014 is used as the test set, and Newstest2013 is used as the validation set. In both datasets, tokens are split into a 32000 word-piece dictionary \citep{wu2016google} which is shared in source and target languages.

\paragraph{Model specifications}

For the WMT14 dataset, we use the default network architecture of the \texttt{base} Transformer model in \citet{vaswani2017attention}, which consists of a 6-layer encoder and 6-layer decoder. The size of hidden nodes and embeddings are set to 512. For the IWSLT14 dataset, we use a smaller architecture, which consists of a 5-layer encoder, and a 5-layer decoder. The size of hidden states and embeddings are set to 256 and the number of heads is set to 4.

\paragraph{Hyperparameter specifications}

Hyperparameters ($\tau, \gamma_\mathit{st}, \gamma_\mathit{tr}, \lambda, \mu$) are determined to make the scales of three loss components similar after initialization. Specifically, we use $\tau=0.3, \gamma_\mathit{st}=0.1, \gamma_\mathit{tr}=0.9,$ $ \lambda=5.0, \mu=1.0$ for IWSLT14 De-En, $\tau=0.3, \gamma_\mathit{st}=0.5, \gamma_\mathit{tr}=0.9, \lambda=5.0, \mu=1.0$ for WMT14 De-En and WMT14 En-De.

\paragraph{BLEU scores}

We use the BLEU score \citep{papineni2002bleu} as our evaluation measure. During inference, we set $C$ to $2, -2, 2$ for WMT14 En-De, De-En and IWSLT14 De-En datasets respectively, according to the average lengths of different languages in the training sets. When using the teacher to rescore, we set $B=4$ and thus have 9 candidates in total. We also evaluate the average per-sentence decoding latencies on one NVIDIA TITAN Xp GPU card by decoding on WMT14 En-De test sets with batch size 1 for our ART teacher model and NART models, and calculate the speedup based on them.

\section{Case Study}
\label{sec:case}
We provide some case studies for the NART models with and without hints in Table~\ref{exp:case-study}. From the first case, we can see that the model without hints translates the meaning of \emph{``as far as I'm concerned''} to a set of meaningless tokens. In the second case, the model without hints omits the phrase \emph{``the farm''} and replaces it with a repetitive phrase \emph{``every morning''}. In the third case, the model without hints mistakenly puts the word \emph{``uploaded''} to the beginning of the sentence, whereas our model correctly translates the source sentence. In all cases, hint-based training helps the NART model to generate better target sentences. 